\documentclass{article}

\usepackage[final]{timeseries_workshop}

\usepackage[utf8]{inputenc} 
\usepackage[T1]{fontenc}    
\usepackage{hyperref}       
\usepackage{url}            
\usepackage{booktabs}       
\usepackage{amsfonts}       
\usepackage{nicefrac}       
\usepackage{microtype}      
\usepackage{xcolor}         

\usepackage{graphicx}
\usepackage{multirow}
\usepackage{enumitem}
\usepackage{wrapfig}
\usepackage{caption}
\usepackage{amsmath}

\title{Frequency Matters: When Time Series Foundation Models Fail Under Spectral Shift}

\author{%
  Tianze Wang\thanks{Tianze Wang (\texttt{tianze.wang@kreditz.com}) completed this work entirely while working at King AI Labs (Microsoft Gaming) and is currently affiliated with Kreditz AB, Stockholm, Sweden.} , 
  Sofiane Ennadir\thanks{Sofiane Ennadir and Gabriela Zarzar Gandler are also affiliated with KTH Royal Institute of Technology.} , 
  John Pertoft, 
  Gabriela Zarzar Gandler\footnotemark[2] , \\[2pt]
  {\bf Lele Cao\thanks{Senior authorship: Sahar Asadi, Oleg Smirnov, and Lele Cao (corresponding, \texttt{lele.cao@king.com}).} , Zineb Senane\thanks{Zineb Senane (\texttt{zineb@fever.energy}) is affliated with Fever Energy, Stockholm, Sweden.} , Styliani Katsarou, Sahar Asadi\footnotemark[3] , Axel Karlsson, Oleg Smirnov\footnotemark[3]} \\[4pt]
  King AI Labs, Microsoft Gaming, Malmskillnadsgatan 19, 111 57 Stockholm, Sweden\\
  \texttt{firstname.lastname@king.com} \\
}

\begin{document}

\maketitle

\begin{abstract}
Time series foundation models (TSFMs) have shown strong results on public benchmarks, prompting comparisons to a ``BERT moment'' for time series. Their effectiveness in industrial settings, however, remains uncertain. We examine why TSFMs often struggle to generalize and highlight spectral shift (a mismatch between the dominant frequency components in downstream tasks and those represented during pretraining) as a key factor. We present evidence from an industrial-scale player engagement prediction task in mobile gaming, where TSFMs underperform domain-adapted baselines. To isolate the mechanism, we design controlled synthetic experiments contrasting signals with seen versus unseen frequency bands, observing systematic degradation under spectral mismatch. These findings position frequency awareness as critical for robust TSFM deployment and motivate new pretraining and evaluation protocols that explicitly account for spectral diversity.
\end{abstract}

\section{Introduction}
Time series data are pervasive across various domains, including finance, healthcare, energy, and gaming. Rapid expansion of time series applications, combined with the success of deep learning in computer vision and natural language processing~\cite{zerveas2021transformer, woo2023moment, zhou2023one}, has led to increased efforts to adapt these models to temporal data. As the volume of time series data continues to grow, manual annotation becomes increasingly expensive and difficult to scale. In response, foundation models trained via self-supervised learning (SSL) have gained attention as a scalable solution. These models are trained on large collections of unlabeled time series using techniques such as contrastive learning~\cite{zhang2022survey}, next-value prediction, and masked autoencoding~\cite{he2022masked}, as well as recently developed joint embedding predictive architectures (JEPA)~\cite{ennadir2024joint}.

However, time series are heterogeneous across domains, as sampling rates, periodicities, and nonstationarities differ significantly between, e.g., electricity, healthcare, finance, and gaming. 
Such diversity is especially pronounced in gaming telemetry, where player behavior exhibits irregular, multi-scale rhythms. This raises a central question: \emph{why do Time Series Foundation Models (TSFMs) that excel on curated public datasets underperform in real gaming applications?}

We present evidence from an industrial-scale Player Engagement Prediction (PEP) task and propose a frequency-based explanation for the TSFM underperformance. Specifically, we hypothesize that TSFMs rely on frequency components memorized during pretraining; when a downstream dataset’s dominant bands fall outside this spectrum, generalization suffers. To probe this, we build controlled synthetic experiments that contrast ``seen'' versus ``unseen'' spectral bands.

\vspace{-3pt}
\section{An Industrial Case: Player Engagement Prediction (PEP) in Gaming}
\vspace{-4pt}
PEP refers to the task of predicting a player's future behavioral and transactional outcomes over a fixed horizon, based on their past interaction history. In this work, we consider a 30-day prediction horizon and use multivariate time series (MTS) data extracted from Candy Crush Saga\footnote{\url{https://www.king.com/game/candycrush}, a AAA mobile match-3 game developed by King.}.
Each MTS sample corresponds to a player’s gameplay sequence over a recent lookback window. We define two labels to be predicted:
\begin{enumerate}[leftmargin=25pt, topsep=0pt, partopsep=0pt, itemsep=0pt]
    \item \textbf{Purchase vs No Purchase}: a binary classification label indicating whether the player is expected to make a purchase within the 30-day horizon.
    \item \textbf{Engagement Score}: a continuous regression target reflecting in-game behavioral intensity (e.g., playtime). Values are normalized to avoid disclosure of business-sensitive information.
\end{enumerate}
\vspace{-4pt}

\subsection{The Industrial Dataset and Evaluation Protocol}
\vspace{-4pt}
Each input sample is an MTS consisting of up to 512 completed gamerounds (time steps) from a single player, captured over a maximum lookback window of 30 days. The input includes 32 univariate time series features, and the missing values are explicitly encoded. Non-exhaustive example features and their categories include:

\noindent
\begin{minipage}[t]{0.63\textwidth}
\begin{itemize}[leftmargin=10pt, topsep=0pt, partopsep=0pt, itemsep=0pt]
    \item \textbf{Progression}: level difficulty, success/fail, attempt count, etc.
    \item \textbf{Gameplay}: time between rounds, number of moves, etc.
    \item \textbf{Resource}: purchase/usage/balance of lives, booster, etc.
    \item \textbf{Strategy}: participation in special game features, etc.
    \item \textbf{Context}: hour of day, days since install, device type, etc.
\end{itemize}
\end{minipage}
\hfill
\begin{minipage}[t]{0.35\textwidth}
    \centering
    \vspace{-20pt}
    \includegraphics[width=\linewidth]{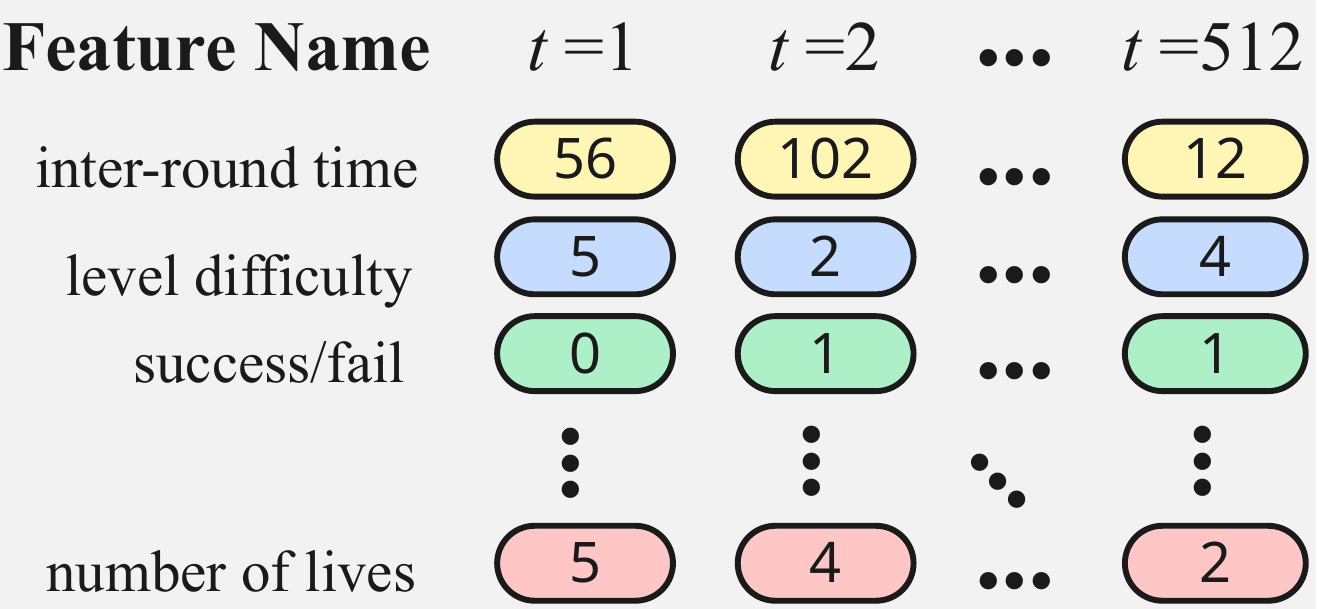}
    \vspace{-15pt}
    \captionof{figure}{\footnotesize An example MTS sample.}
    \label{fig:mts-sample}
\end{minipage}

All features are scaled using min-max or log transformations with application-specific capping, and categorical fields are encoded numerically when applicable. Commercially sensitive financial signals are normalized and anonymized where present. Player-identifiable data are not used in our dataset.

We evaluate models on two complementary sets: 
(i) a \textbf{player-holdout set}, where train, validation, and test splits are created on disjoint groups of players without temporal separation, ensuring no per-user leakage while largely preserving stationarity; and 
(ii) a \textbf{temporal-holdout set}, where models are trained and validated on earlier periods and then evaluated \emph{zero-shot} on the future evaluation set, with no tuning of model weights or hyper-parameters.
The primary industrial dataset spans 226 days and contains 824,208 MTS samples from approximately 40,000 selected players, with validation and test (a.k.a., player-holdout) sets of 178,279 and 176,632 samples. The temporal-holdout set covers a 28-day period about two months after the end of the primary dataset window, comprising 34,279 samples from roughly 7,000 players. All features are standardized per-feature using training statistics.

\vspace{-6pt}
\subsection{The Industrial Benchmarking}
\vspace{-4pt}
We evaluate (i) industrial baselines (XGBoost \cite{chen2016xgboost}, TabNet\cite{arik2021tabnet}) on temporally aggregated features, (ii) PatchTST \cite{zhang2022patchtst} as a strong task fine-tuned model \cite{zhang2022patchtst}, and (iii) TSFMs that include a representative open model (MOMENT-small) \cite{woo2023moment}. 
TSFMs are assessed under linear probing with lightweight heads. Unless noted, {\bf temporal-holdout} results are zero-shot on the temporal-holdout set, with baselines trained on the primary data directly evaluated on that holdout month.

We report \textbf{Accuracy/AUC} for the purchase classification label and \textbf{MSE/MAE} for the engagement-intensity regression label. 
Table~\ref{tab:engagement} summarizes the results. Despite public-benchmark success, the TSFM lags behind the domain-adapted PatchTST and tabular baselines in this industrial setting.

\begin{table}[h]
\centering
\caption{Experimental results for Player Engagement Prediction (PEP).}
\label{tab:engagement}
\small
\setlength{\tabcolsep}{4pt}
\renewcommand{\arraystretch}{1.15}
\begin{tabular}{l|cc|cc|cc|cc}
\bottomrule
\multirow{3.5}{*}{Model} & \multicolumn{2}{c}{Accuracy $\uparrow$} & \multicolumn{2}{c}{AUC $\uparrow$} & \multicolumn{2}{c}{MSE $\downarrow$} & \multicolumn{2}{c}{MAE $\downarrow$} \\
\cmidrule(lr){2-3}\cmidrule(lr){4-5}\cmidrule(lr){6-7}\cmidrule(lr){8-9}
& {\footnotesize \shortstack{player\\holdout}}  & {\footnotesize \shortstack{temporal\\holdout}}
& {\footnotesize \shortstack{player\\holdout}} & {\footnotesize \shortstack{temporal\\holdout}}
& {\footnotesize \shortstack{player\\holdout}} & {\footnotesize \shortstack{temporal\\holdout}}
& {\footnotesize \shortstack{player\\holdout}} & {\footnotesize \shortstack{temporal\\holdout}} \\
\midrule
XGBoost$^{*}$ & 0.841 & 0.801 & 0.915 & 0.883 & 1.200 & 1.310 & 0.780 & 0.850 \\
TabNet$^{*}$  & 0.836 & 0.795 & 0.911 & 0.852 & 1.304 & 2.140 & 0.852 & 1.169 \\
PatchTST & \textbf{0.939} & \textbf{0.921} & \textbf{0.982} & \textbf{0.975} & \textbf{0.518} & 0.711 & \textbf{0.489} & \textbf{0.586} \\
MOMENT-small & 0.758 & 0.701 & 0.791 & 0.749 & 2.250 & 2.854 & 1.151 & 1.324 \\
\toprule
\end{tabular}
\vspace{-10pt}
\end{table}

\vspace{-2pt}
\section{A Frequency Perspective}\label{sec:freq_perspective}
\vspace{-3pt}
As discussed in the previous section, we observe that foundation models for time series, such as MOMENT, underperform compared to traditional fully supervised baselines when considering datasets from real-world use cases. From a spectral analysis of the set of frequencies present in the considered dataset and the pretraining dataset, we observe the existence of a gap (the complete study is provided in Appendix \ref{app:spectral_analysis}). We consequently hypothesize that this performance gap stems, at least in part, from a fundamental mismatch between the statistical properties of the pretraining data and those of the downstream tasks. In contrast to text-based foundation models (e.g., large language models), where linguistic structure and semantics exhibit strong shared patterns across domains, time series data are highly sensitive to changes in frequency, resolution, and temporal dynamics. A minor shift in dominant frequency bands can result in drastically different signal characteristics, undermining the generalization capacity of pretrained models. 

Consequently, while LLMs have had some success in the generalization aspect of text-related tasks, where structural and semantic consistency support robust transfer, time series data often lack such stability. Domain-specific variations, such as sampling rate and seasonality, can lead to significant distribution shifts that current Time Series-based Foundation Models are not able to handle. While other factors, such as covariate shift, nonstationarity, and label misalignment, also play important roles, we argue that the potential frequency shift is a primary driver of transfer degradation.
\begin{quote}
\textit{\textbf{Main Hypothesis:}} Time Series Foundation Models (TSFMs) transfer effectively when the downstream series share dominant frequency bands with those represented during pretraining; performance degrades significantly otherwise.
\end{quote}
To validate this hypothesis, we empirically investigate how shifts in frequency affect the downstream performance of a TSFM. We design a controlled setup in which we contrast the model’s performance on data with \textit{seen} versus \textit{unseen} frequency bands. Specifically, starting from a dataset used during pretraining, we construct two derived datasets: one that preserves the dominant frequency components from pretraining, and another with altered spectral characteristics outside the pretrained distribution.

By design, the foundation model has been exposed to the frequency profile of the first dataset, while the second represents a spectral domain it has not encountered. Poor generalization to the latter would suggest that the model has not truly learned generalizable temporal representations, but rather memorized patterns associated with specific frequency bands.

\subsection{Experimental Setup.}

While we provide an exhaustive description of the data generation process in Appendix \ref{app:data_generation}, we highlight the main steps taken in the following section.
Given a chosen time series dataset that was used for the pretraining, the data generation process can be summarized as follows:

\begin{enumerate}[leftmargin=1.5em]
    \item \textbf{Frequency extraction:} For each series, we compute the FFT and retain the top-$5$ dominant frequencies. Let $[f^{\text{low}}, f^{\text{high}}]$ denote the resulting band of interest.
    
    \item \textbf{Signal generation:} We synthesize two sets of signals: \emph{Seen} band, where frequencies are sampled uniformly from $[f^{\text{low}}, f^{\text{high}}]$; and \emph{Unseen} band, where frequencies are sampled from $[f^{\text{low}}+\delta,\, f^{\text{high}}+\delta]$. The shift $\delta$ is chosen such that the two intervals are disjoint and remain within the overall frequency range of the original series. Each signal is constructed as a sum of sinusoids with random phases and light additive noise.
    
    \item \textbf{Labeling:} Regression targets are the $z$-score normalized sum of sampled frequencies to remove scale effects; classification labels indicate whether a sample comes from the seen or unseen band.
    
    \item \textbf{Evaluation:} We freeze a pretrained TSFM backbone and train lightweight regression or classification heads on each synthetic variant.
\end{enumerate}

Following the generation process, we consider MOMENT \cite{woo2023moment} as the basis TSFM for our experiment. We used a linear head to produce the final prediction, which was trained during the downstream task. We based our experiments on a set of classification datasets that were used to train the model. Specifically, we considered time series representing sensor outputs (FordA, FordB \cite{dau2019ucr} and FaultDetectionA and FaultDetectionB \cite{lessmeier2016condition}), we additionally considered time series representing consumers' electricity behavior (SmallKitchenAppliances and ElectricDevices \cite{lines2011electricitydataset}). For all the experiments, the models were trained using the Adam optimizer, binary cross-entropy loss (for classification) and mean-square error loss (for regression). All experiments were run $3$ times, and we report the average and standard deviation to ensure a robust analysis and to reduce the effect of randomness. Additional details about our implementation are provided in Appendix \ref{app:experimental_details}.

\subsection{Experimental Results}
Table~\ref{tab:synthetic} reports the mean (and corresponding standard deviation) Mean Squared Error (MSE) and Mean Absolute Error (MAE) of the seen and unseen generated datasets. As expected, we see that in all cases, the resulting MSE and MAE for the seen dataset are smaller than those from the unseen datasets. We note that in these experiments, the model remains frozen while only a regression head is trained; therefore, we are testing the model's ability to extract meaningful representations of the time series, which could be useful for the downstream task. From these results, we can conclude that the MOMENT model has a better ability to deal with datasets with similar semantic characteristics (such as frequencies) to those used for training.

\begin{table}[h]
\vspace{-10pt}
\centering
\caption{Regression performance of a frozen TSFM encoder (MOMENT-small) with a trainable regressor on synthetic datasets.}
\label{tab:synthetic}
\small
\setlength{\tabcolsep}{6pt}
\renewcommand{\arraystretch}{1}
\begin{tabular}{l|cc|cc}
\toprule
\multirow{2}{*}{Dataset} & \multicolumn{2}{c|}{Test MSE} & \multicolumn{2}{c}{Test MAE} \\
 & Seen (\checkmark) & Unseen ($\times$) & Seen (\checkmark) & Unseen ($\times$) \\
\midrule
FordA & $0.333 \pm 0.010$ & $0.366 \pm 0.005$ & $0.439 \pm 0.005$ & $0.457 \pm 0.005$ \\
FordB & $0.333 \pm 0.010$ & $0.358 \pm 0.008$ & $0.426 \pm 0.005$ & $0.456 \pm 0.005$ \\
ElectricDevices & $0.644 \pm 0.002$ & $0.952 \pm 0.003$ & $0.559 \pm 0.001$ & $0.791 \pm 0.004$ \\
SmallKitchenAppliances & $0.691 \pm 0.059$ & $0.877 \pm 0.017$ & $0.686 \pm 0.031$ & $0.752 \pm 0.007$ \\
FaultDetectionA & $0.689 \pm 0.001$ & $0.942 \pm 0.004$ & $0.666 \pm 0.001$ & $0.779 \pm 0.001$ \\
FaultDetectionB & $1.129 \pm 0.172$ & $2.005 \pm 0.266$ & $0.875 \pm 0.084$ & $1.140 \pm 0.034$ \\
\bottomrule
\end{tabular}
\end{table}

In addition to the previously considered regression task, we also extended the results and the study to include a classification downstream task, where similar trends are observed, further validating our hypothesis. The results of the analysis are provided in Appendix \ref{app:additional_results}.

\subsection{Limitations and Practical Implications}

In this section, we would like to declare a few limitations of this study. 
Our evidence is drawn from one industrial domain (mobile gaming) and a single TSFM configuration, and we are conducting broader validations at the moment. The synthetic probes also simplify real-world dynamics, relying on sinusoidal signals that do not fully capture irregular sampling, burstiness, or regime shifts. Despite these constraints, the consistent trends suggest practical guidance: practitioners should assess spectral overlap between downstream data and pretraining corpora, apply frequency-aware augmentations or light adaptation when overlap is low, and adopt benchmarks that explicitly stress-test robustness under spectral shifts. We share more detailed discussions about frequency-aware pretraining in Appendix~\ref{app:pretraining_reflection}.

\section{Conclusion}

Most recent works on TSFMs report strong results on carefully curated public benchmarks. However, the statistical properties of these datasets are far from those encountered in messy, real-world applications.
Benchmark success does not guarantee industrial transfer. Using an industrial-scale gaming task and controlled experiments, we trace TSFM underperformance to spectral shift between pretraining and downstream signals. Because frequency composition varies sharply across domains, sampling regimes, and tasks, addressing spectral shift is essential for TSFMs to move from benchmark success towards universality. We recommend (i) quantifying spectral overlap between pretraining corpora and downstream datasets, (ii) incorporating frequency-aware augmentations and pretraining strategies, and (iii) adopting benchmarks that explicitly probe robustness to spectral diversity. 

\bibliographystyle{plain}
\bibliography{refs}

\newpage

\appendix

\vbox{%
\hsize\textwidth
\linewidth\hsize
\vskip 0.1in
\centering
{\LARGE\bf Supplementary Material of\\ Frequency Matters: When Time Series Foundation Models Fail Under Spectral Shift\par}
\vspace{2\baselineskip}
}

\section{Related Work}\label{app:related_work}

Following the success of foundation models in computer vision and natural language processing (NLP), and with the rise of time series as a valid modality within the broader context of deep learning, a growing number of foundation models have been proposed to tackle time series. While different downstream applications are possible, the majority of these models have focused on forecasting. Some approaches, such as GPT4TS \cite{zhou2023one} and Time-LLM \cite{jin2024timellm}, adapt existing large language models by ``reprogramming'' them for the temporal domain, freezing most layers while finetuning lightweight, time series–specific modules. In contrast, a wave of recent work has introduced models trained entirely from scratch on large-scale, heterogeneous time series corpora.

Lag-Llama \cite{rasul2023lagllama} presents a general-purpose foundation model for univariate probabilistic forecasting, built on a decoder-only transformer. By explicitly modeling lagged dependencies, it adapts autoregressive language modeling techniques to temporal patterns, achieving strong performance on long-horizon forecasting tasks, though it is primarily designed for forecasting rather than broader time series applications.

Chronos \cite{ansari2024chronos} adopts a probabilistic formulation by discretizing continuous time series into tokens and applying autoregressive sequence modeling. Released as a family of five models ranging from 20M to 710M parameters, Chronos allows for flexible trade-offs between efficiency and accuracy. While forecasting remains central, its probabilistic generation framework also enables extensions to classification and anomaly detection via reinterpretation of generated trajectories.

Moirai \cite{pmlr-v235-woo24a} introduces a scalable temporal foundation model with hierarchical attention tailored to long and irregular sequences. Through multi-resolution temporal representations, Moirai can capture dependencies across a wide range of timescales, making it effective not only for forecasting but also for classification and imputation tasks across domains such as healthcare and finance.

TimesFM \cite{das2024decoder} builds a general-purpose forecaster trained on a large collection of public and proprietary datasets. With a transformer-based encoder and decoder backbone, it demonstrates strong zero-shot performance across diverse application domains, including finance, energy, and traffic. Beyond forecasting, TimesFM produces robust embeddings that transfer effectively to downstream classification and anomaly detection tasks, particularly in low-label regimes.

MOMENT \cite{woo2023moment} proposes a universal time series foundation model that unifies forecasting with representation learning across multiple modalities. Combining large-scale pretraining with adaptive fine-tuning, MOMENT supports a broad suite of tasks, including forecasting, classification, anomaly detection, and imputation. Its architecture is explicitly designed for multimodality, enabling the integration of time series with contextual signals such as categorical features or text.

In terms of training methodology, these models share the same underlying philosophy of large-scale pretraining on heterogeneous time series data, but diverge in their design choices. Lag-Llama, Chronos, and TimesFM employ transformer-based encoders with objectives such as next-step prediction and masked modeling; Chronos, in particular, leverages tokenization to enable language-model style autoregression. On the other hand, Moirai emphasizes architectural scalability, using hierarchical attention and multi-resolution sampling to pretrain on very long sequences. MOMENT combines forecasting objectives with contrastive pretraining, producing general-purpose temporal embeddings that are transferable across tasks.

Overall, a diverse set of methodologies and architectural choices has been proposed to address time series downstream tasks ranging from forecasting to classification and anomaly detection. While these models demonstrate strong performance on widely used benchmark datasets, their evaluation remains confined mainly to controlled settings. In contrast, our work emphasizes real-world applications, focusing on practical use cases that reflect the challenges and constraints encountered in industrial deployment scenarios, as detailed in the main paper.

\section{Spectral Analysis}\label{app:spectral_analysis}

\begin{figure}[h]
    \centering
    \includegraphics[width=\linewidth]{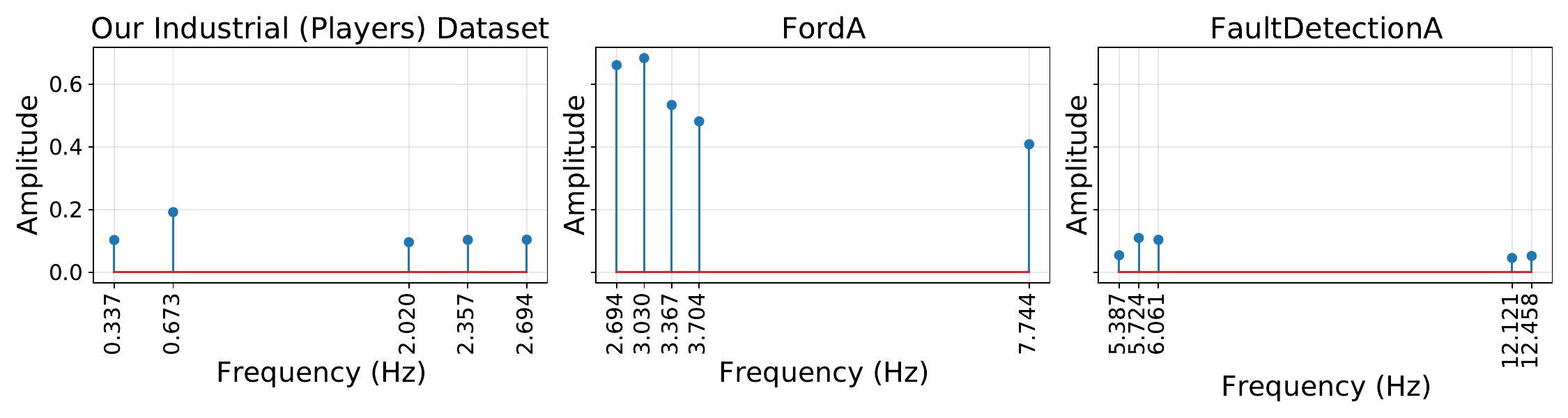}
    \caption{Analysis of the dominant frequencies in our dataset and in the datasets used for pretraining MOMENT (FordA and FaultDetectionA).}
    \label{fig:frequency_analysis}
\end{figure}

To better understand why the time series foundation model MOMENT underperforms on our dataset, we analyze its dominant frequency components and compare them to those of the datasets used during pretraining. Figure \ref{fig:frequency_analysis} illustrates this comparison, focusing on the FordA and FaultDetectionA datasets from the UCR time series repository, which were part of MOMENT’s pretraining corpus.

As shown in the figure, our dataset exhibits fundamentally different dominant frequencies compared to those observed during pretraining. This discrepancy suggests that MOMENT was not exposed to such frequency patterns before, which likely contributed to its poor generalization and performance.

This analysis highlights the potential importance of frequency alignment in the generalization capability of TSFMs. It also motivates the frequency-based hypothesis we explore further in Section \ref{sec:freq_perspective}.

\section{Data Generation}\label{app:data_generation}
In the following section, and as a complement to Section \ref{sec:freq_perspective}, we provide additional details about the synthetic data generation process. 

As previously explained, we constructed \emph{seen} and \emph{unseen} synthetic datasets from a set of considered datasets that were used to pretrain the considered foundation model. Specifically, for a real sequence $x(t)$, we compute its discrete Fourier transform (DFT) and retain the top-$k$ dominant components,
\begin{equation}
    \mathcal{F}\{x(t)\} \Rightarrow F(x) = \{(f_i,a_i)\}_{i=1}^k,
\end{equation}

where $f_i$ and $a_i$ denote dominant frequencies and amplitudes. Based on the previous frequency extraction, synthetic signals are generated by recombining these components with controlled perturbations (additive Gaussian noise and bounded amplitude scaling). This yields:

\begin{itemize}
    \item \textbf{Seen samples:} dominant content constrained within dataset-level frequency bounds:
        \begin{equation*}
            f_i \in [f_{\text{sim}}^{\text{low}}, f_{\text{sim}}^{\text{high}}],
        \end{equation*}    
    with only mild perturbations;
    \item \textbf{Unseen samples:} at least one dominant component outside the empirical bounds,
    \begin{equation*}
       f_j \notin [f_{\text{sim}}^{\text{low}} + \delta, f_{\text{sim}}^{\text{high}}  + \delta], 
    \end{equation*}
    or taken from complementary/shifted bands (low-only, high-only, or randomized). For instance, if the dominant range is $[0,20]$ Hz, then the unseen dataset would be generated from $[20+\delta,\,40+\delta]$ with $\delta$ being drawn at random subject to the constraint that the out-of-band upper limit does not exceed the maximum frequency present in the original series.
\end{itemize}

We note that the dataset-wide frequency bounds $[f_{\text{sim}}^{\text{low}}, f_{\text{sim}}^{\text{high}}]$ are estimated from the distribution of dominant frequencies across the dataset (e.g., via quantile summaries or min/max).

Based on the generated seen and unseen datasets, we generate a set of train/val/test splits for each dataset, obtained with a fixed $70/15/15$ partition. Each split contains single-channel sequences of fixed length $L$ (typically $L=512$),
\begin{equation}
    X^{(s)} \in \mathbb{R}^{N_s \times 1 \times L}, \qquad s \in \{\text{train},\text{val},\text{test}\}.
\end{equation}

\textbf{Regression targets.} Given the previously generated time series, we need to construct labels that could be used in the downstream task. In this perspective, each synthetic sample is paired with a continuous regression target derived directly from its generation metadata. Let $\mathcal{F}_{\text{used}} = \{ f_j \}$ denote the set of frequencies used in the synthesis. The scalar target is defined as the sum of frequencies:
\begin{equation}
    y = \sum_{f_j \in \mathcal{F}_{\text{used}}} f_j.
\end{equation}

We note that while the previous labeling operation provides a frequency-aware label, it does indeed naturally increase with spectral displacement and distinguishes between seen and unseen samples. 

Since such displacement can be used by the model to make the task easier, we use a normalization mechanism to standardize the labels and make them comparable in terms of scale:
\begin{equation}
    \tilde{y}^{(s)} = \frac{y^{(s)} - \mu_y}{\sigma_y}, 
    \qquad 
    y = \sigma_y \,\tilde{y}^{(s)} + \mu_y,
\end{equation}

After this normalization operation, the labels for both seen and unseen datasets are within the same range, and therefore, we would expect the model to rather use temporal representations to extract the label.

\begin{figure}[h]
    \centering
    \includegraphics[width=.9\linewidth]{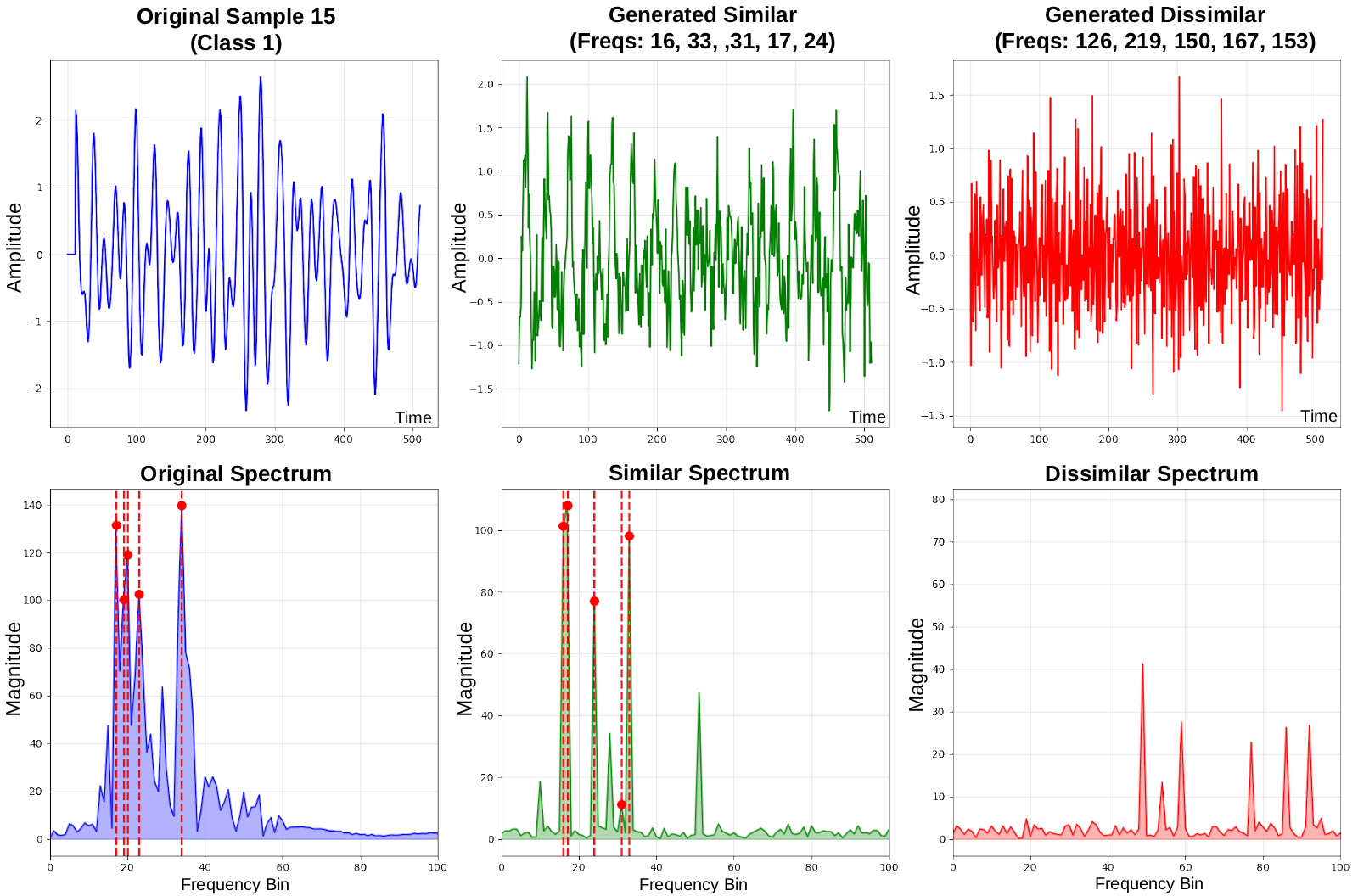}
    \caption{Synthetic series from seen vs. unseen frequency bands. Top: time domain; Bottom: spectra.}
    \label{fig:freq}
\end{figure}

In Figure \ref{fig:freq}, we present an example of the generation, where given an original sample, we analyze its original spectrum to generate the seen and unseen datasets. 

\section{Additional Results - Classification} \label{app:additional_results}

To further illustrate our hypothesis beyond the regression task, we considered a classification task. Specifically, we bin adjacent frequency ranges (e.g., $[0, 10]$ vs. $[10, 20]$ Hz) and train a binary classifier on top of the frozen encoder. The binning threshold is chosen as the median of the frequency ranges from the train split to ensure a balanced class distribution.

\begin{table}[h]
\centering
\caption{Classification performance of a frozen TSFM encoder (MOMENT-small) with a trainable binary classifier on synthetic datasets.}
\label{tab:synthetic_classification}
\small
\setlength{\tabcolsep}{6pt}
\renewcommand{\arraystretch}{1.05}
\begin{tabular}{l|cc|cc}
\toprule
\multirow{2}{*}{Dataset} & \multicolumn{2}{c|}{Test Accuracy} & \multicolumn{2}{c}{Test AUC} \\
 & Seen (\checkmark) & Unseen ($\times$) & Seen (\checkmark) & Unseen ($\times$) \\
\midrule
FordA & $0.837 \pm 0.004$ & $0.829 \pm 0.001$ & $0.926 \pm 0.002$ & $0.901 \pm 0.001$ \\
FordB & $0.826 \pm 0.000$ & $0.838 \pm 0.002$ & $0.903 \pm 0.002$ & $0.926 \pm 0.001$ \\
ElectricDevices & $0.785 \pm 0.003$ & $0.650 \pm 0.003$ & $0.890 \pm 0.001$ & $0.716 \pm 0.002$ \\
SmallKitchenAppliances & $0.708 \pm 0.010$ & $0.756 \pm 0.041$ & $0.833 \pm 0.018$ & $0.859 \pm 0.007$ \\
FaultDetectionA & $0.766 \pm 0.003$ & $0.651 \pm 0.003$ & $0.858 \pm 0.000$ & $0.707 \pm 0.003$ \\
FaultDetectionB & $0.444 \pm 0.048$ & $0.472 \pm 0.096$ & $0.519 \pm 0.306$ & $0.476 \pm 0.190$ \\
ECG5000 & $0.809 \pm 0.020$ & $0.591 \pm 0.028$ & $0.898 \pm 0.007$ & $0.692 \pm 0.007$ \\
SwedishLeaf & $0.689 \pm 0.020$ & $0.600 \pm 0.040$ & $0.796 \pm 0.021$ & $0.643 \pm 0.028$ \\
\bottomrule
\end{tabular}
\end{table}

Table \ref{tab:synthetic_classification} provides the mean and corresponding standard deviation of the Test accuracy and the Test AUC. As previously seen in the regression task, the performance is high for ranges within the presumed pretraining spectrum but drops for out-of-range comparisons (e.g., $[40, 50]$ vs. $[50, 60]$ Hz), reinforcing the frequency-mismatch explanation.

\section{Experimental Details} \label{app:experimental_details}
In all experiments, we use linear probing: the MOMENT backbone (i.e.,~the patch embedder and transformer encoder) is frozen, and only the regression or classification head is trained. The model is optimized using the Adam optimizer (learning rate $10^{-3}$) for 50 epochs, with a mean squared error loss (for regression) or binary cross-entropy loss (for classification). We use the same hidden dimensionality as the one from the pretrained chosen backbone, corresponding to the predefined MOMENT configurations. Model selection is performed based on validation MSE, and the best checkpoint is then evaluated on the test set. We report all metrics in Table~\ref{tab:engagement}.

We use a combination of CPU, NVIDIA GPUs (L4 24GB, T4 16GB), and an Apple MPS device (M1 MAX 32GB) to conduct the experiments in this paper. In the player engagement use case, we use CPU for the XGBoost model, a single GPU for TabNet, and PatchTST. We utilize distributed data parallelism with up to eight GPUs to accelerate the training process for the MOMENT model on the large mobile gaming dataset. In the frequency-perspective experiments, we use a single GPU and an Apple MPS device for the regression tasks.

\section{Reflection on the Role of Frequencies in TSFM Pretraining} \label{app:pretraining_reflection}

In this work, we report an empirical observation about TSFMs: a pretrained TSFM can underperform on domain-specific downstream tasks relative to models trained directly on in-domain time series. To probe this effect, we empirically evaluate how well a pretrained TSFM performs on synthetically generated time series whose dominant frequencies are varied. We observe a trend in frequency-alignment effect: the TSFM performs better on synthetic time series whose dominant frequencies overlap with those seen during pretraining than on synthetic time series samples from unseen or underrepresented frequency bands.

Our results suggest that, beyond standard self-supervised objectives for times-series pretraining (e.g., masked modeling, forecasting, and contrastive learning), the spectral composition of the pretraining data may also play a critical role in the model's generalization and robustness. One might consider adding an auxiliary task during pretraining to predict the dominant frequency bands or adopting frequency-based data augmentation or sampling to expand frequency coverage. We hypothesize that frequency-aware pretraining strategies, combined with standard time series pretraining methods, may be a fruitful future direction for improving the performance of TSFMs in zero-shot/few-shot settings.

\end{document}